\documentclass[journal]{IEEEtran}
\usepackage[T1]{fontenc}
\usepackage{graphicx}
\usepackage{times}
\usepackage{helvet}
\usepackage{courier}
\usepackage{amsmath}
\usepackage{algorithm}
\usepackage{algorithmic}
\usepackage{csquotes} 
\usepackage{color}
\usepackage{paralist}
\usepackage{amssymb}
\usepackage{indentfirst}
\usepackage{subfigure}
\usepackage{float}
\usepackage{multirow}
\usepackage{cite}
\usepackage{mathrsfs}

\usepackage{mathrsfs} 
\usepackage{amsfonts}
\usepackage{todonotes}
\usepackage{pgfplots} 
\usepackage{epstopdf}
\usepackage{epsfig}
\pgfplotsset{compat=newest}
\usepackage{color}
\usepackage{enumitem}
\usepackage{CJKutf8}
\definecolor{forestgreen}{RGB}{0,139,69}

\usepackage{xcolor}
\definecolor{citecolor}{HTML}{0071bc}
\usepackage[colorlinks, linkcolor=red,  anchorcolor=blue, citecolor=citecolor]{hyperref} 

\usepackage{xcolor}
\definecolor{SeaGreen4}{RGB}{0,205,102} 
\definecolor{SlateBlue}{RGB}{106,90,205} 
\definecolor{DarkRed}{RGB}{178,34,34} 
	
\usepackage[switch]{lineno}
\usepackage{float}
\usepackage{textcomp,booktabs}
\usepackage{amssymb}
\usepackage{pifont}

\usepackage{makecell}

\usepackage{colortbl}
\definecolor{mygray}{gray}{.9}
\definecolor{mypink}{rgb}{.99,.91,.95}
\definecolor{mycyan}{cmyk}{.3,0,0,0}

\begin{document}

\title{
ESTR-CoT: Towards Explainable and Accurate Event Stream based Scene Text Recognition with Chain-of-Thought Reasoning
}

\author{Xiao Wang, \emph{Member, IEEE}, Jingtao Jiang, Qiang Chen, Lan Chen*, Lin Zhu, 
    \\ Yaowei Wang, \emph{Member, IEEE}, Yonghong Tian, \emph{Fellow, IEEE}, Jin Tang

\thanks{$\bullet$ Xiao Wang, Jingtao Jiang, Qiang Chen, Jin Tang are with the School of Computer Science and Technology, Anhui University, Hefei 230601, China. (email: \{xiaowang, tangjin\}@ahu.edu.cn, e23301220@stu.ahu.edu.cn, jingtaoj16@gmail.com)} 

\thanks{$\bullet$ Lan Chen is with the School of Electronic and Information Engineering, Anhui University, Hefei 230601, China. (email: chenlan@ahu.edu.cn)}

\thanks{$\bullet$ Lin Zhu is with Beijing Institute of Technology, Beijing, China. (email: linzhu@pku.edu.cn)} 

\thanks{$\bullet$ Yaowei Wang is with Harbin Institute of Technology, Shenzhen, China; Peng Cheng Laboratory, Shenzhen, China. (email: wangyw@pcl.ac.cn)} 

\thanks{$\bullet$ Yonghong Tian is with Peng Cheng Laboratory, Shenzhen, China; School of Computer Science, Peking University, China; School of Electronic and Computer Engineering, Shenzhen Graduate School, Peking University, China (email: yhtian@pku.edu.cn)}

\thanks{* Corresponding Author: Lan Chen (chenlan@ahu.edu.cn)}  
}

\markboth{ IEEE Transactions on ***, 2025 } 
{Shell \MakeLowercase{\textit{et al.}}: Bare Demo of IEEEtran.cls for IEEE Journals}

\maketitle

\begin{abstract}
Event stream based scene text recognition is a newly arising research topic in recent years which performs better than the widely used RGB cameras in extremely challenging scenarios, especially the low illumination, fast motion. Existing works either adopt end-to-end encoder-decoder framework or large language models for enhanced recognition, however, they are still limited by the challenges of insufficient interpretability and weak contextual logical reasoning. 
In this work, we propose a novel chain-of-thought reasoning based event stream scene text recognition framework, termed ESTR-CoT. Specifically, we first adopt the vision encoder EVA-CLIP (ViT-G/14) to transform the input event stream into tokens and utilize a Llama tokenizer to encode the given generation prompt. A Q-former is used to align the vision token to the pre-trained large language model Vicuna-7B and output both the answer and chain-of-thought (CoT) reasoning process simultaneously. Our framework can be optimized using supervised fine-tuning in an end-to-end manner. 
In addition, we also propose a large-scale CoT dataset to train our framework via a three stage processing (i.e., generation, polish, and expert verification). This dataset provides a solid data foundation for the development of subsequent reasoning-based large models. 
Extensive experiments on three event stream STR benchmark datasets (i.e., EventSTR, WordArt*, IC15*) fully validated the effectiveness and interpretability of our proposed framework. 
The source code and pre-trained models will be released on \url{https://github.com/Event-AHU/ESTR-CoT}. 
\end{abstract}


\IEEEpeerreviewmaketitle

\section{Introduction}

\begin{figure*}[t]
\centering
\includegraphics[width=0.98\linewidth]{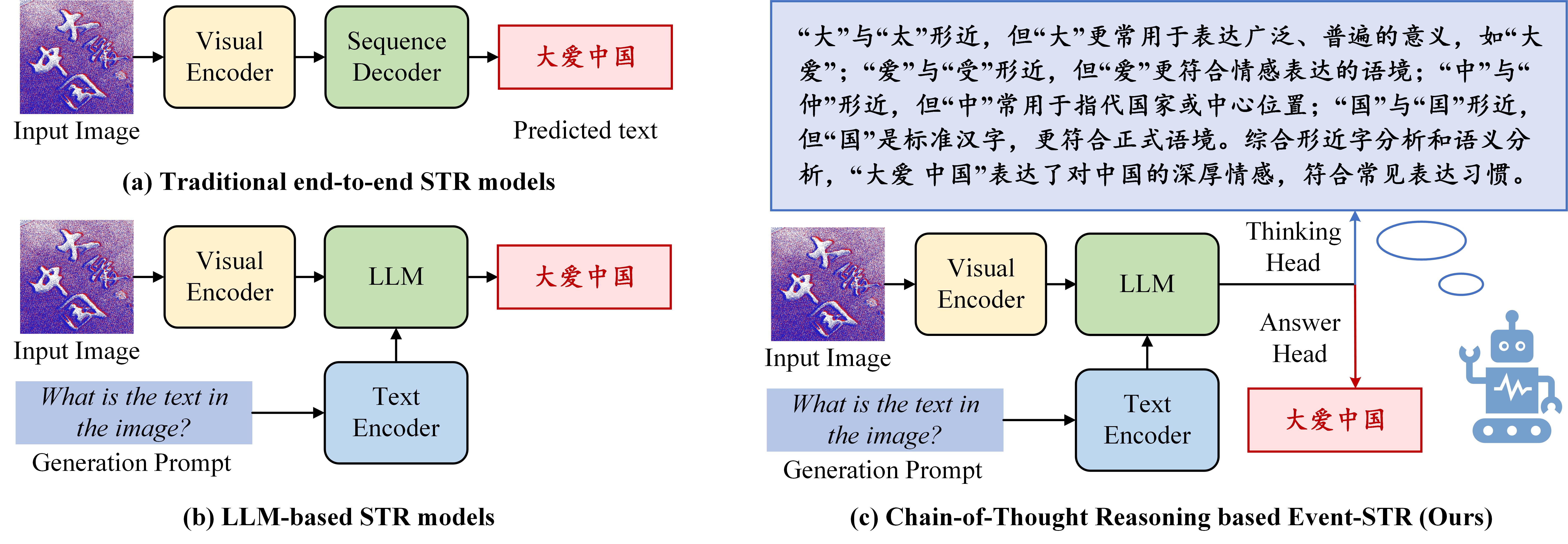}
\caption{\textbf{
Comparison between the (a) traditional end-to-end encoder-decoder STR, (b) LLM-based STR, and (c) our newly proposed chain-of-thought reasoning based event stream STR.} Traditional STR models rely on task-specific architectures and direct visual-text mappings. LLM-based approaches introduce language understanding but still lack explicit reasoning. Our framework incorporates chain-of-thought reasoning to enable interpretable and logically grounded recognition.}
\label{fig:firstIMG}
\end{figure*}

\IEEEPARstart{S}{cene} Text Recognition (STR) targets to understand and recognize the words in the given scene using a machine learning model. It has been widely exploited based on the RGB frame cameras and achieves significant improvements with the help of deep neural networks. This task can be applied in autonomous driving, augmented reality, document digitization, retail and e-commerce, etc. However, the performance of STR in challenging scenarios remains unsatisfactory due to the use of RGB cameras, such as fast motion and extreme illumination. Therefore, there is still a long way to go in the research of robust STR models.

Recently, Large Language Models (LLMs) have achieved great success in the Natural Language Processing (NLP) community, such as ChatGPT~\cite{achiam2023gpt}, GPT-4o~\cite{hurst2024gpt}, DeepSeek~\cite{bi2024deepseek}, Qwen~\cite{bai2023qwen}, etc. The LLMs are also introduced into the multi-modal scenarios and can be applied to visual question answering~\cite{antol2015vqa,alayrac2022flamingo,li2023blip,hu2024bliva}, medical report generation~\cite{jing2017automatic,cao2023mmtn,chen2024medblip,Wang_2025_CVPR}, etc. Some researchers also adopt the LLMs for the scene text recognition~\cite{liu2024textmonkey, feng2023docpedia, wei2025vary, hu2024mplug, wei2024OCR2.0, wang2025eventstr} and achieve significant improvements over non-LLMs based STR models, as shown in Fig.~\ref{fig:firstIMG} (a, b).  
More in detail, TextMonkey~\cite{liu2024textmonkey} proposes a high-resolution, location-aware LLM to unify OCR and VQA tasks; DocPedia~\cite{feng2023docpedia} directly processes frequency-domain document images without relying on traditional OCR; and Vary~\cite{wei2025vary} extends visual vocabulary and enhances multilingual document understanding, highlighting the potential of LLMs in fine-grained text recognition. 
Some researchers resort to event cameras for the perception in extremely challenging scenarios (i.e., low illumination, fast motion) to replace or assist RGB cameras, e.g., event-based object detection~\cite{gehrig2023RVTs,Guo2024SpatiotemporalAggregationTransformer,peng2024scene,wang2025object,wang2025CvHeat}, tracking~\cite{zhang2023AFNet,Chen_2019,zhang2021object,wang2024eventvot}, and scene text recognition~\cite{wang2025eventstr}. EventSTR~\cite{wang2025eventstr} proposed by Wang et al. introduces a large-scale benchmark dataset for event-based text recognition and proposes SimC-ESTR, a framework that combines event based vision with large language models through vision-text alignment, memory-enhanced reasoning, and glyph-level correction, achieving superior robustness under challenging visual conditions.

Despite these breakthroughs, we can find that these scene text recognition models are still limited by the following issues: 
1). Mainstream STR algorithms typically use RGB frames as input, making them susceptible to challenges such as low illumination, high-speed motion, and overexposure. 
2). Mainstream STR algorithms lack interpretability and strong reasoning abilities, even when adopting the Large Language Models. It is difficult to explicitly model relationships between different characters, however, the context-based reasoning is key to achieving high-performance recognition in challenging scenarios. 
3). The reasoning ability relies on the release of large-scale, high-quality chain-of-thought datasets. However, there is still no dataset in academia specifically for event stream based scene text recognition, which further limits the reasoning capabilities of existing models. 
Therefore, it is natural to raise the following question: ``\textit{how to design an event-stream based scene text recognition framework that can effectively integrate contextual information to enable high-quality reasoning, leveraging the capabilities of large language models to achieve more accurate recognition?}"

To address the aforementioned issues, in this paper, we propose a novel chain-of-thought reasoning based event stream scene text recognition framework, termed ESTR-CoT. As shown in Fig.~\ref{fig:firstIMG} (c), the framework we propose is built upon a large language model, with the core innovation being the introduction of a reasoning task head, which enables high-quality contextual reasoning while simultaneously achieving accurate text recognition. Specifically, given the input event streams, we first adopt a vision encoder (EVA-CLIP (ViT-G/14)~\cite{sun2023eva} adopted in this paper) to transform them into event tokens and use a Llama-Tokenizer~\cite{touvron2023llama} to encode the given generation prompt. We concatenate these tokens with the randomly initialized query tokens and feed them into a Q-former network to align the vision tokens for the large language models. Then, we adopt the pre-trained large language model Vicuna-7B~\cite{chiang2023vicuna} to decode the \texttt{thinking} and \texttt{answer} simultaneously. 
To bridge the data gaps and achieve the training of our framework, as shown in Fig.~\ref{fig:CoTdataGeneration}, we resort to the large language model DeepSeek-V3 to generate the chain-of-thought reasoning data based on the ground truth answers. The additional evaluation on the generated data by the LLMs and human experts is also needed for high-quality CoT data generation. Some representative samples are provided in Fig.~\ref{fig:dataset}, and an overview of our proposed framework can be found in Fig.~\ref{fig:framework}.


To sum up, the main contributions of this paper can be summarized as the following three aspects: 

1). We propose a novel event stream scene text recognition framework based on chain-of-thought reasoning, termed ESTR-COT. The algorithm we propose can significantly enhance the interpretability of scene text recognition, improving recognition accuracy through contextual reasoning. 

2). We build a large-scale reasoning dataset for the event stream based scene text recognition. It contains 16,222 image-reasoning pairs, each consisting of an event-based scene text image, the corresponding recognized text (answer), and a detailed CoT rationale explaining how the model arrives at the final answer.

3). Extensive experiments on three widely used event-based STR benchmark datasets (i.e., EventSTR~\cite{wang2025eventstr}, WordArt*~\cite{xie2022CornerTransformer}, IC15*~\cite{karatzas2015icdar}) fully validated the effectiveness and interpretability of our proposed reasoning strategy.

\textit{The rest of this paper is organized as follows:} 
We review the related works on the scene text recognition, large language model and reasoning models in section~\ref{sec::relatedWorks}. In section~\ref{sec::methods}, we introduce our proposed reasoning strategy for the large language model based scene text recognition. In section~\ref{sec::experiments}, we conduct the experimental analysis and provide both quantitative and qualitative analyses. Finally, in section~\ref{sec::conclusion}, we conclude this paper and propose possible research directions on this work.

\section{Related Works} \label{sec::relatedWorks}

In this section, we introduce the related works from Scene Text Recognition, LLM and Reasoning Models, Event-based Vision. More details can be found in the following surveys~\cite{wang2023MMPTMsurvey, huang2023reasoningLLMsurvey, long2021SceneTextDetRec} and paper list~\footnote{\url{https://github.com/Event-AHU/OCR_Paper_List}}. 

\subsection{Scene Text Recognition} 
Scene text recognition~\cite{wang2011end,shi2016end,liao2019scene,han2024spotlight} naturally involves both vision and language processing. Traditional research efforts often emphasized either visual feature extraction or language modeling, but recent advances have sought a more balanced integration of both modalities to enhance robustness across diverse scenarios. E2STR~\cite{zhao2024E2STR} enhances adaptability by introducing context-rich text sequences and a context training strategy, offering improved flexibility across environments. CCD~\cite{Guan2023CCD} leverages a self-supervised segmentation module and character-to-character distillation to improve text representation learning, while SIGA~\cite{guan2023SIGA} further refines segmentation via implicit attention alignment. CDistNet~\cite{zheng2024cdistnet} incorporates both visual and semantic positional embeddings into its transformer-based design to address irregular text layouts and complex backgrounds.

In parallel, several works introduce iterative error correction strategies using language models. VOLTER~\cite{li2024volter}, BUSNet~\cite{Wei2024BUSNet}, MATRNet~\cite{na2022matrn}, LevOCR~\cite{da2022levocr}, and ABINet~\cite{fang2021ABINet} exemplify this trend, integrating language models to refine recognition through feedback loops, improving both robustness and interoperability. Building on this foundation, recent research has moved toward large language model (LLM)-based scene text recognition. These models exploit the generative and contextual reasoning capabilities of LLMs to unify visual and linguistic understanding. For instance, TextMonkey~\cite{liu2024textmonkey} is a multimodal LLM optimized for text-centric tasks, utilizing high-resolution inputs and location-aware responses to support enhanced interaction and interpretability. DocPedia~\cite{feng2023docpedia} eliminates the need for traditional OCR by processing high-resolution document images in the frequency domain, efficiently capturing both visual and textual cues. Vary~\cite{wei2025vary} enriches the visual vocabulary of vision-language models, enabling fine-grained document OCR and chart understanding, especially in multilingual contexts. mPLUG-DocOwl 1.5~\cite{hu2024mplug} introduces Unified Structure Learning for improved document layout comprehension, while OCR2.0~\cite{wei2024OCR2.0} presents a powerful 580M-parameter model capable of handling OCR tasks ranging from text recognition to formula parsing. Despite these advances, LLM-based models still struggle under extreme conditions such as low lighting, blur, or complex noise. These limitations highlight the ongoing challenge of achieving robust, generalizable scene text recognition across varied and adverse environments. To address these challenges, EventSTR~\cite{wang2025eventstr} pioneers the use of event cameras for scene text recognition. 


\subsection{LLM and Reasoning Models}
Large Language Models (LLMs) are pretrained models exhibiting strong capabilities in natural language understanding and generation. They serve as the foundational backbone for many advanced reasoning techniques by enabling coherent and contextually relevant text generation. Notable examples include the GPT series from OpenAI~\cite{radford2019language,brown2020language}, Google's PaLM~\cite{chowdhery2023palm}, Meta's LLaMA~\cite{touvron2023llama}, as well as instruction-tuned variants such as InstructGPT~\cite{ouyang2022training} and Vicuna~\cite{chiang2023vicuna}. Although these models differ in scale and training data, they share a common versatility, being adaptable to a broad spectrum of downstream tasks, including reasoning, dialogue systems, and code generation.

Building upon LLMs, reasoning models~\cite{Xue2025EVQA} explicitly structure the inference process to enhance interpretability and accuracy. One effective technique is Chain of Thought (CoT) prompting, which encourages the model to generate intermediate, step-by-step rationales instead of only final answers~\cite{kojima2022large}. This method substantially improves performance on complex reasoning tasks by rendering the model's decision-making process more transparent. Further extending CoT, frameworks such as Tree-of-Thoughts and Graph-of-Thoughts facilitate the parallel exploration and integration of multiple reasoning paths, thereby increasing robustness, especially in scenarios involving ambiguity or multiple valid solutions~\cite{yao2023tree, besta2024topologies}. To reduce noise and hallucinations inherent in single-chain reasoning, self-consistency approaches aggregate outputs from multiple independent reasoning chains~\cite{wang2022self}. Additionally, reinforcement learning techniques optimize reasoning processes by refining reward functions that emphasize coherence, verifiability, and task-specific criteria, exemplified by models like LLaVA-Reasoner~\cite{zhang2024improve}. Complementing these advances, the integration of vision-language foundation models such as BLIP-2 and MiniGPT-4 provides powerful multimodal encoders, enabling reasoning pipelines to extend beyond text to encompass images, audio, and other modalities~\cite{li2023blip, zhu2023minigpt}. Collectively, these developments constitute a versatile and comprehensive toolkit for eliciting interpretable, high-quality reasoning across diverse application domains.

\subsection{Event-based Vision}  
Event cameras~\cite{gallego2020event, jiang2024evcslr,jiang2023masked} are bio‑inspired sensors that asynchronously record per‑pixel brightness changes with microsecond‑level latency, offering high temporal resolution, low power consumption, and wide dynamic range. These properties make event cameras well‑suited for tasks in fast‑moving or poorly lit environments, including autonomous driving, robotic perception, and medical imaging. 

In human activity recognition, ESTF~\cite{wang2024hardvs} projects raw event streams into learned spatial–temporal embeddings, enabling robust classification under rapid motion and low illumination. For visual tracking, EventVOT~\cite{wang2024eventvot} provides a large‑scale (1280$\times$720) event‑only dataset of 1,141 sequences spanning pedestrians, vehicles, drones, and sports objects, alongside a hierarchical distillation framework that yields high‑speed, low‑latency tracking. Recurrent Vision Transformers (RVTs)~\cite{gehrig2023RVTs} exploit event cameras’ temporal fidelity to achieve accurate object detection in dynamic scenes. SAFE~\cite{li2023semantic} fuses event streams with RGB frames and semantic labels via a pretrained vision‑language backbone, bridging the modality gap and overcoming the limitations of small‑scale networks. Despite these advances, event‑based methods have seen little application in scene text recognition. To fill this gap, EventSTR~\cite{wang2025eventstr} introduces a novel task and dataset for event‑stream‑based STR, comprising 9,928 high‑definition (1280$\times$720) event sequences annotated with both Chinese and English text under diverse lighting, motion, and occlusion conditions. 


\section{Our Proposed Approach} \label{sec::methods}

\subsection{Overview}
\begin{figure*}[t]
\centering
\includegraphics[width=\linewidth]{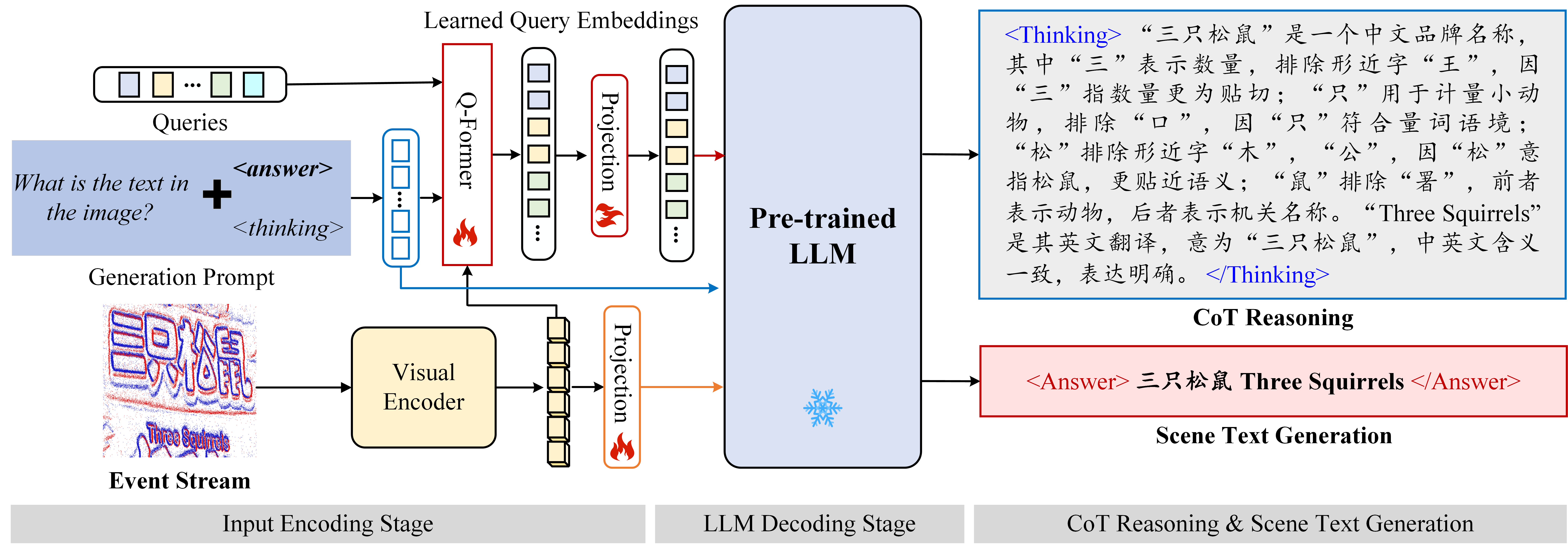}
\caption{\textbf{An overview of the proposed ESTR-CoT framework for the reasoning-based event stream scene text recognition.} This figure illustrates the prompt-based control architecture for generating answers and reasoning chains from event-based scene text inputs. A Q-Former extracts learned query embeddings from visual features captured by the event stream. These embeddings are fused with a fixed textual prompt and passed to a pre-trained LLM. By appending different suffixes (\texttt{\textcolor{magenta}{<answer>}} or \texttt{\textcolor{blue}{<thinking>}}) to the prompt, the model is guided to generate either the final recognition result or the corresponding reasoning chain.
}
\label{fig:framework}
\end{figure*}

In this paper, we propose ESTR-CoT, a reasoning-enhanced framework designed to improve both the accuracy and interpretability of scene text recognition in visually challenging environments. The core idea is to enable the model to produce not only the correct answer but also a coherent reasoning process, thereby improving transparency and interoperability. As shown in Fig.~\ref{fig:framework}, ESTR-CoT consists of the following four core components, i.e., visual encoder, Q-Former, pre-trained LLM, and prompt-based control architecture. Specifically, the visual encoder is a pre-trained EVA-CLIP (ViT-G/14)~\cite{sun2023eva} which is used to extract discriminative visual features from event-based scene images. Q-Former aligns visual features with a tokenized textual prompt to generate query embeddings. These embeddings act as a bridge between the visual encoder and the language model, capturing the multi-modal context. A large language model is the foundation of our framework, which receives the projected visual features and prompt embeddings to auto-regressively generate textual outputs. It produces either the final recognition result or the corresponding reasoning chain. For the prompt-based control architecture, different prompt suffixes (\texttt{\textcolor{magenta}{<answer>}} and \texttt{\textcolor{blue}{<thinking>}}) are used to steer the generation target. During training, the model is supervised with both answer annotations and reasoning chains from high-quality CoT data. A multi-task loss is applied to jointly optimize both objectives. At inference time, ESTR-CoT can generate only the final answer for efficiency or include the reasoning chain for better transparency.


\subsection{Chain-of-Thought Data Generation}

\begin{figure*}
\centering
\includegraphics[width=0.98\linewidth]{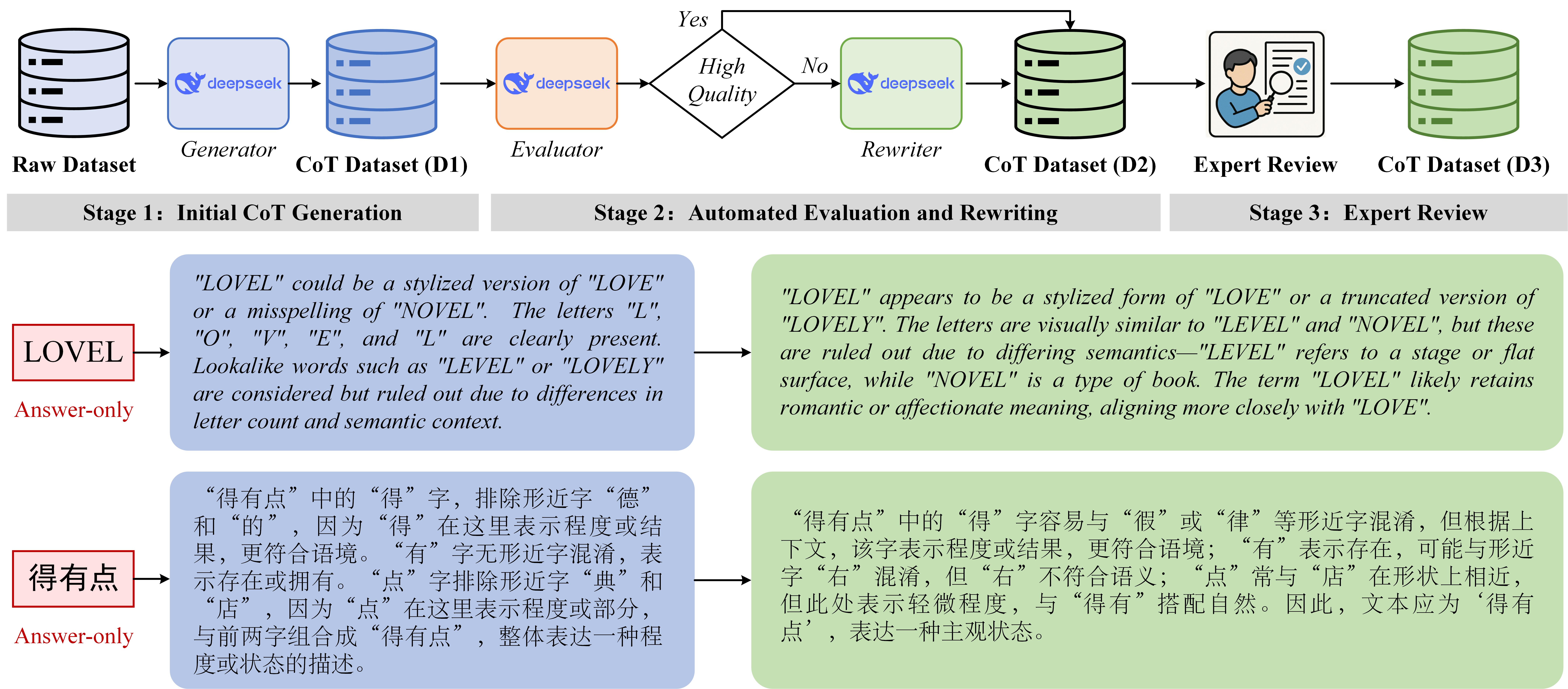}
\caption{\textbf{Data Generation Pipeline of ESTR-CoT.}
The figure illustrates a three-stage pipeline for constructing high-quality CoT data in ESTR-CoT. Beginning with a raw dataset, a generator produces initial CoT responses (D1). An evaluator filters these outputs: high-quality samples proceed to D2, while suboptimal ones are revised by a rewriter based on evaluator feedback. All refined samples are added to D2. Finally, human experts review D2 to curate the final expert-approved dataset D3. The bottom example shows how reasoning is refined from a basic explanation (Stage 1) to a more precise and semantically coherent version (Stage 2).} 
\label{fig:CoTdataGeneration}
\end{figure*}

To enhance the reasoning capability of large vision-language models in the context of scene text recognition, we construct a high-quality Chain-of-Thought dataset via a structured multi-stage pipeline, as illustrated in Fig.~\ref{fig:CoTdataGeneration}. Unlike conventional text recognition annotations that only provide flat textual labels, our CoT corpus incorporates fine-grained visual-semantic reasoning, enabling models to explicitly consider ambiguities such as lookalike characters, semantic context, and domain-specific constraints.

\noindent $\bullet$ \textbf{Stage 1: Initial CoT Generation.}  
Given a raw dataset of OCR labels or question-answer pairs (e.g., the text string ``LOVEL'' predicted from an image), we first utilize a powerful LLM (e.g., Deepseek-V3) to generate an initial CoT explanation. The LLM is prompted with task-specific instructions that require it to analyze both visual similarity (e.g., ``LOVE'' vs. ``NOVEL'') and semantic coherence. The resulting outputs are structured in the format:
\begin{quote}
\texttt{\textcolor{magenta}{<answer>}LOVEL\textcolor{magenta}{</answer>}\textcolor{blue}{<thinking>}"LOVEL" could be a stylized version of "LOVE" or a misspelling of "NOVEL". The letters "L", "O", "V", "E", and "L" are clearly present.  Lookalike words such as "LEVEL" or "LOVELY" are considered but ruled out due to differences in letter count and semantic context.\textcolor{blue}{</thinking>}}
\end{quote}
This produces the first-stage dataset $D_1$, which consists of \texttt{(input text, reasoning)} pairs.

\noindent $\bullet$ \textbf{Stage 2: Automatic Evaluation and Rewriting.}  
To guarantee logical validity, informativeness, and structural clarity of the generated reasoning, we employ an automatic evaluation module that examines each CoT explanation against the following criteria:
\begin{enumerate}[label=(\arabic*)]
\item \textit{Length Constraint:} reasoning chains exceeding a preset maximum token length (e.g., 100 tokens) are flagged for rewriting, preventing overly verbose or unfocused explanations;
\item \textit{Visual-Semantic Completeness:} the explanation must explicitly include both visual form analysis (e.g., character shape comparisons) and semantic context reasoning (e.g., word plausibility), ensuring comprehensive reasoning coverage;
\item \textit{Logical Consistency:} the chain must maintain coherent argumentative flow without contradictions or irrelevant content.
\end{enumerate}

Samples passing all checks are directly added to the curated dataset $D_2$. Failed cases are sent to an LLM-powered rewriter module for improved reformulation. The rewritten explanations undergo re-evaluation and are included in $D_2$ only upon satisfying the quality standards. This iterative process effectively filters out noisy or superficial chains of thought while enhancing reasoning quality.

\noindent $\bullet$ \textbf{Stage 3: Expert Review and Validation.}  
While the automatic modules are highly effective, some complex or ambiguous cases require human judgment. Therefore, samples in $D_2$ are reviewed by human experts with background in scene text understanding. The experts assess correctness, clarity, and linguistic fluency. Only the most reliable samples are retained to construct the final dataset $D_3$, which serves as the training source for CoT-aware STR models.

\begin{figure*}[t]
\centering
\includegraphics[width=\linewidth]{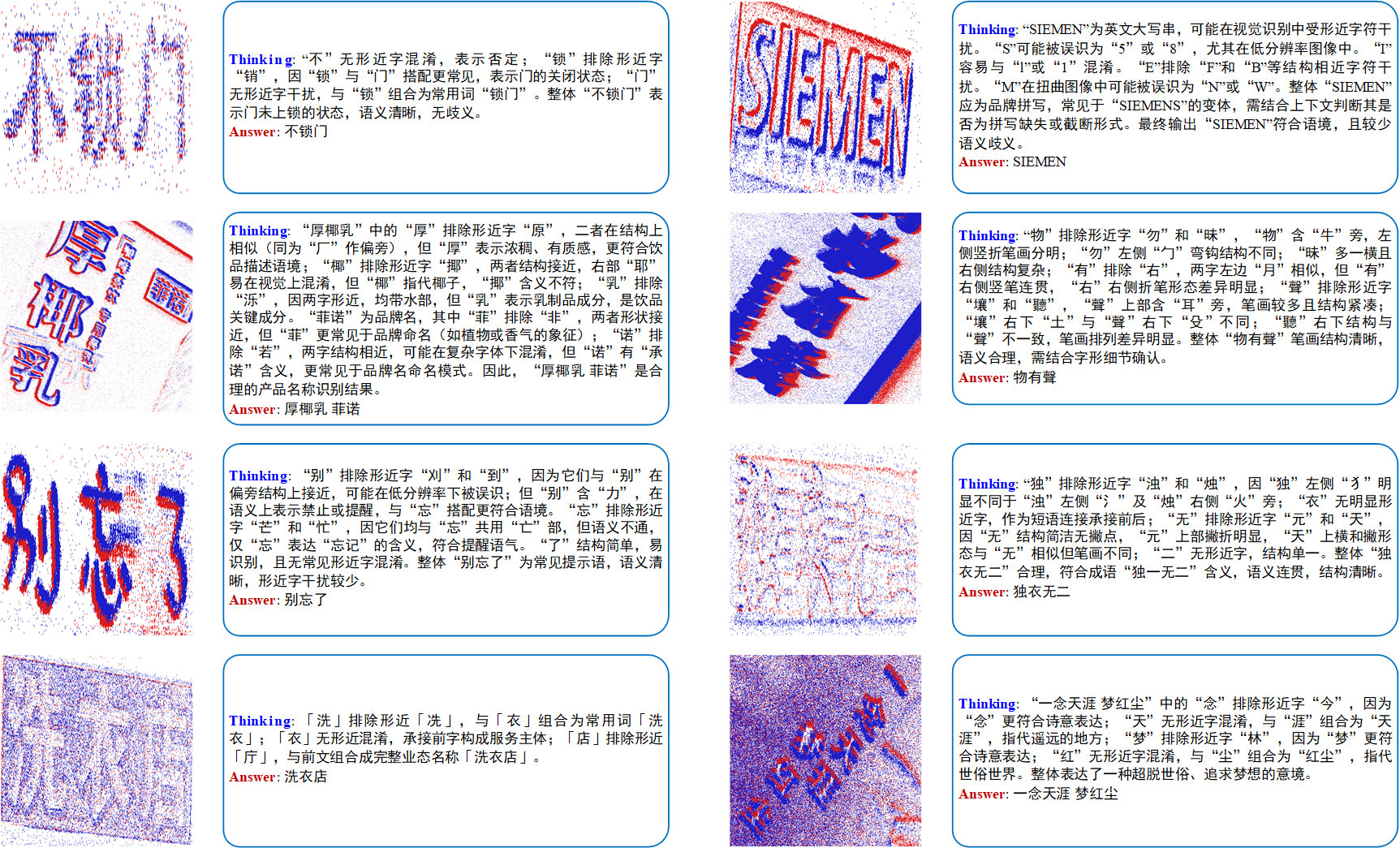}
\caption{\textbf{Representative samples from our newly proposed CoT\_ESTR Dataset.} 
Each example consists of an event stream scene image, a recognition result \texttt{\textcolor{magenta}{<Answer>}}, and an accompanying reasoning chain \texttt{\textcolor{blue}{<Thinking>}}.}  
\label{fig:dataset}
\end{figure*}

Consider the OCR output ``LOVEL'', which can be ambiguously interpreted. Initially, the CoT reasoning explores multiple candidate words: it could be a stylized version of ``LOVE'' or a misspelling/truncation related to ``NOVEL''. The letters ``L'', ``O'', ``V'', ``E'', and ``L'' are clearly present, while lookalike words such as ``LEVEL'' or ``LOVELY'' are also considered but ruled out due to inconsistencies in letter count or semantic context.

Further analysis focuses on the visual similarity of individual letters and their semantic alignment. Although ``LOVEL'' shares visual traits with ``LEVEL'' and ``NOVEL'', the latter candidates are discarded based on their differing meanings—``LEVEL'' denotes a stage or flat surface, and ``NOVEL'' refers to a type of book, which do not align with the romantic or affectionate context suggested by the imagery. Consequently, the reasoning favors ``LOVE'' as the most plausible interpretation, viewing ``LOVEL'' as a stylized or truncated variant emphasizing romantic connotations.

As shown in Fig.~\ref{fig:CoTdataGeneration}, this case exemplifies how the Stage 2 automatic evaluation and rewriting module refines ambiguous initial explanations by emphasizing letter shape details and semantic context, resulting in a more precise, semantically coherent, and visually robust Chain-of-Thought.

\textbf{Benefits.}  
This pipeline ensures that the final CoT dataset not only captures domain-specific reasoning patterns but also minimizes annotation noise. It supports multiple downstream tasks, including OCR error correction, few-shot generalization, and instruction-tuned STR modeling. By incorporating both machine-filtered and human-verified samples, we achieve a balance between scalability and quality, which is essential for training reliable reasoning-augmented vision models. 
Formally, our pipeline constructs a reasoning-augmented dataset through the following process:
\begin{align}
D_1 &= \{ (a_i, c_i^{(0)}) \}_{i=1}^{N} \\
D_2 &= \{ (a_i, c_i^{(1)}) \mid \text{Eval}(c_i^{(0)}) = \text{Pass or Revised} \} \\
D_3 &= \{ (a_i, c_i^{*}) \mid (a_i, c_i^{(1)}) \in D_2, \ \text{HumanPass}(c_i^{(1)}) = \text{True} \}
\end{align}
Here, $a_i$ denotes the original OCR output or answer, $c_i^{(0)}$ is the initial CoT generated by the LLM, $c_i^{(1)}$ is the reasoning after automatic evaluation and possible rewriting, and $c_i^{*}$ is the final expert-approved reasoning.

The automatic evaluation function $\text{Eval}(\cdot)$ checks each reasoning chain $c$ as:
\begin{align}
\text{Eval}(c) =
\begin{cases}
\text{Pass}, & \text{if } \text{Len}(c) < L_{\max} \\
             & \quad \wedge~\text{HasVisual}(c) \\
             & \quad \wedge~\text{HasSemantic}(c) \\
\text{Fail}, & \text{otherwise}
\end{cases}
\end{align}
where $L_{\max}$ is the maximum allowed length, and the predicates $\text{HasVisual}(\cdot)$ and $\text{HasSemantic}(\cdot)$ ensure the reasoning contains both visual and semantic analysis. 

Each final sample in $D_3$ is formatted as:
\begin{equation}
\begin{aligned}
x_i = {} & \text{\texttt{\textcolor{magenta}{<answer>}}} a_i \text{\texttt{\textcolor{magenta}{</answer>}}} \\
         & \text{\texttt{\textcolor{blue}{<thinking>}}} c_i^{*} \text{\texttt{\textcolor{blue}{</thinking>}}}
\end{aligned}
\end{equation}
which allows straightforward downstream usage in training CoT-aware models. An illustration of the dataset structure and examples is shown in Fig.~\ref{fig:dataset}.

\subsection{Model Design}

After acquiring high-quality Chain-of-Thought data \(D_3\) via the previously proposed data generation pipeline, we leverage this reasoning-enriched dataset to improve the model's reasoning ability. As illustrated in Fig.~\ref{fig:framework}, we first explain the motivation for producing two separate outputs \texttt{answer} and \texttt{think}, and then outline two architectural variants developed to effectively model the interaction between visual and textual modalities.

\noindent $\bullet$ \textbf{Input Construction. } 
Given an event-based scene image $\mathcal{I} \in \mathbb{R}^{C \times H \times W}$, we use a pre-trained visual encoder, EVA-CLIP~\cite{sun2023eva} (ViT-G/14), to extract discriminative visual features. The encoder divides the image $\mathcal{I}$ into fixed-size patches (14$\times$14 px), flattens them into tokens, and applies multi-head self-attention to obtain the visual feature map $F_v$. A global [CLS] token provides a holistic representation of the image. To combine the visual features with textual information, we introduce a Q-Former module that aligns $F_v$ with tokenized prompt embeddings $F_\ell$ to produce query embeddings $F_q$. The textual prompt is set to:
\[
\mathcal{P} = \text{\small\texttt{"What is the text in the image?"}},
\]
which is then tokenized and encoded into embeddings $F_\ell$. The Q-Former processes the visual features $F_v$ and prompt embeddings $F_\ell$ to generate the query embeddings $F_q$.

The final model input is constructed by combining the visual and textual components:
\begin{equation}
\mathrm{Input}_{\mathrm{LLM}} = \left[\mathrm{Proj}(F_q), \; \mathrm{Proj}(F_v), \; F_\ell\right]
\end{equation}

where \(F_\ell\) is the tokenized prompt embedding, \(\mathrm{Proj}(F_q)\) is the projection of query embeddings, and  \(\mathrm{Proj}(F_v)\) is the projection of the visual feature map. This input is used for both generating the answer and the reasoning chain in the subsequent architecture.

\noindent $\bullet$ \textbf{Why Separate Outputs for \texttt{answer} and \texttt{thinking}?} 
The decision to output both \texttt{answer} and \texttt{thinking} arises from the need to balance high recognition accuracy with interpretability in complex scene text recognition tasks.

\textit{1). Improved Interpretability:}  
By outputting \texttt{thinking}, the model provides not only the correct answer but also articulates the reasoning behind its decision. This transparency helps users understand the model's decision-making process, which is especially valuable in domains like document analysis or autonomous systems where reasoning is crucial.

\textit{2). Enhanced Model Robustness:}  
Separate outputs allow the model to treat answer generation and reasoning as distinct tasks. This enables better management of the interplay between visual features and textual context. The \texttt{thinking} output ensures that the answer is contextually grounded and logically sound.

\textit{3). Multi-task Learning Synergy:}  
By training the model to predict both the answer and its reasoning, the model learns to optimize the reasoning and answering process simultaneously. This dual-head setup helps refine the model’s understanding, improving both the accuracy and coherence of the output.

\noindent $\bullet$ \textbf{Prompt-based Control Architecture.} 
We adopt a prompt-based control approach to guide the model's generation behavior. Specifically, we use a shared prompt prefix for both the final text output (\texttt{answer}) and the reasoning chain (\texttt{thinking}), distinguishing the two by appending different suffixes:
\begin{equation}
\mathcal{P}_{\text{answer}} = \mathcal{P} \; \textcolor{magenta}{\texttt{<answer>}}, \quad 
\mathcal{P}_{\text{thinking}} = \mathcal{P} \; \textcolor{blue}{\texttt{<thinking>}}
\end{equation}

These prompts are tokenized and encoded into embeddings:
\begin{equation}
F_{\ell_{\text{answer}}} = \mathrm{Encode}(\mathcal{P}_{\text{answer}}), \quad 
F_{\ell_{\text{thinking}}} = \mathrm{Encode}(\mathcal{P}_{\text{thinking}})
\label{eq:encode_prompt}
\end{equation}


The model input for both tasks is formed by concatenating the projected query embeddings \(\mathrm{Proj}(F_q)\), projected visual features \(\mathrm{Proj}(F_v)\), and the corresponding prompt embeddings:
\begin{align}
\mathrm{Input}_{\mathrm{LLM, answer}} &= \left[\mathrm{Proj}(F_q), \mathrm{Proj}(F_v), F_{\ell_{\text{answer}}}\right], \label{eq:llm_input_answer} \\
\mathrm{Input}_{\mathrm{LLM, thinking}} &= \left[\mathrm{Proj}(F_q), \mathrm{Proj}(F_v), F_{\ell_{\text{thinking}}}\right]. \label{eq:llm_input_thinking}
\end{align}


The model then autoregressively generates the answer and reasoning outputs respectively:
\begin{align}
\hat{y}_{\text{answer}} &= \mathrm{LLM}(\mathrm{Input}_{\mathrm{LLM, answer}}), \label{eq:gen_answer} \\
\hat{y}_{\text{thinking}} &= \mathrm{LLM}(\mathrm{Input}_{\mathrm{LLM, thinking}}). \label{eq:gen_thinking}
\end{align}


\noindent $\bullet$ \textbf{Projection-separated Control Architecture.} 
We also explore an alternative architecture that employs separate projection layers for visual and query embeddings for the answer and reasoning tasks, respectively. This variant aims to specialize feature processing for each output by using distinct projections:
\begin{align}
\mathrm{Input}_{\mathrm{LLM, answer}} &= \left[\mathrm{Proj}_{\text{ans}}(F_q), \mathrm{Proj}_{\text{ans}}(F_v), F_{\ell}\right], \label{eq:proj_input_answer} \\
\mathrm{Input}_{\mathrm{LLM, thinking}} &= \left[\mathrm{Proj}_{\text{thinking}}(F_q), \mathrm{Proj}_{\text{thinking}}(F_v), F_{\ell}\right]. \label{eq:proj_input_thinking}
\end{align}

The resulting outputs \(\hat{y}_{\text{answer}}\) and \(\hat{y}_{\text{thinking}}\) are generated in the same autoregressive manner as above. In this work, we primarily focus on the prompt-based control architecture due to its simplicity and efficiency. The projection-separated variant is briefly introduced here and further analyzed in the experiments section~\ref{sec:architecture_analysis}.

\subsection{Head and Loss Function}

To train the model to generate both the final answer and its corresponding reasoning process, we define the total loss as the sum of two components:
\begin{equation}
\mathcal{L} = \mathcal{L}_{\text{answer}} + \mathcal{L}_{\text{thinking}},
\end{equation}
where $\mathcal{L}_{\text{answer}}$ is the cross-entropy loss for answer prediction, and $\mathcal{L}_{\text{thinking}}$ is the loss for generating the reasoning chain. Specifically,
\begin{equation}
\mathcal{L}_{\text{answer}} = \mathrm{CE}\left(\hat{y}_j^{\text{answer}},\ y_j\right),
\end{equation}
\begin{equation}
\mathcal{L}_{\text{thinking}} = \mathrm{CE}\left(\hat{y}_j^{\text{thinking}},\ c_j\right),
\end{equation}
where $\hat{y}_j^{\text{answer}}$ denotes the predicted answer and $y_j$ is the corresponding ground-truth label. Similarly, $\hat{y}_j^{\text{thinking}}$ is the generated reasoning chain, and $c_j$ is the ground-truth chain-of-thought. Although we explored weighted combinations of the two loss components in our experiments (see Section~\ref{sec:Loss_analysis}), we found that the simple summation strategy performs comparably and is more stable. Therefore, we adopt the unweighted sum as the default objective for training.

\section{Experiments} \label{sec::experiments}

\begin{table*}
\centering
\small 
\caption{Comparison of BLEU scores with SOTA methods on the EventSTR dataset.}
\label{tab:bleu}
\resizebox{\textwidth}{!}{ 
\begin{tabular}{l|l|c|cccccc}
\hline 
\textbf{Algorithm}  & \textbf{Publish}  & \textbf{Backbone}  & \textbf{BLEU-1} & \textbf{BLEU-2} & \textbf{BLEU-3} & \textbf{BLEU-4}  &\textbf{Params(M)}& \textbf{Code}   \\ \hline
CCD~\cite{Guan2023CCD}  & ICCV 2023  & ViT  &  0.365 & 0.254 & 0.172 & 0.145 &52.0& \href{https://github.com/TongkunGuan/CCD}{URL}\\
SIGA~\cite{guan2023SIGA}  & CVPR 2023  & ResNet&  0.434& 0.393& 0.346& 0.307 &40.4& \href{https://github.com/TongkunGuan/SIGA}{URL}\\
CDistNet~\cite{zheng2024cdistnet}  & IJCV 2023  & ResNet+Transformer&  0.333 & 0.242 & 0.157 & 0.135 &65.5& \href{https://github.com/simplify23/CDistNet?tab=readme-ov-file}{URL}  \\
PARSeq~\cite{bautista2022PARseq}  & ECCV 2022  & ViT  &  0.450& 0.357& 0.281& 0.224 &23.4& \href{https://github.com/baudm/parseq}{URL}\\
MGP-STR~\cite{wang2022mgpstr}& ECCV 2022& Transformer& 0.427& 0.339& 0.278& 0.232 &148.0&\href{https://github.com/AlibabaResearch/AdvancedLiterateMachinery/tree/main/OCR/MGP-STR}{URL}\\
GOT-OCR2.0~\cite{wei2024OCR2.0}& arXiv 2024 &ViT & 0.426 & 0.390 & 0.358 & 0.332  &580.0&\href{https://github.com/Ucas-HaoranWei/GOT-OCR2.0}{URL}\\  
BLIVA~\cite{hu2024bliva}&  AAAI 2024 & ViT  &0.584 & 0.528 & 0.450 & 0.386 &7531.3& \href{https://github.com/mlpc-ucsd/BLIVA}{URL}  \\
 SimC-ESTR~\cite{wang2025eventstr} &arXiv 2025 & ViT  & 0.638& 0.583 & 0.500 & 0.430 & 7531.3&\href{https://github.com/Event-AHU/EventSTR}{URL} \\
\hline
ESTR-CoT (Ours) &- & ViT  & \textbf{0.648}& \textbf{0.586}& \textbf{0.500}& \textbf{0.430}& 7531.3 &\href{https://github.com/Event-AHU/ESTR-CoT}{URL} \\ 
\hline
\end{tabular}} 
\end{table*}

\begin{table*}
\centering 
\small 
\caption{The accuracy comparisons with SOTA methods on WordArt* and IC15*.}
\label{tab:acc}
\begin{tabular}{l|l|c|c|c|c|c}
\hline 
\multirow{2}{*}{\textbf{Algorithm}}  & \multirow{2}{*}{\textbf{Publish}}  & \multirow{2}{*}{\textbf{Backbone}}  & \multicolumn{2}{c|}{\textbf{Accuracy}} &  \multirow{2}{*}{\textbf{Params(M)}}&\multirow{2}{*}{\textbf{Code}} \\
\cline{4-5}  
&  &  & \textbf{WordArt*}& \textbf{IC15*}&   &\\
\hline 
LISTER~\cite{cheng2023lister}& ICCV 2023& CNN& 55.3& 69.0 &  49.9&\href{https://github.com/AlibabaResearch/AdvancedLiterateMachinery/tree/main/OCR/LISTER}{URL}\\
CCD~\cite{Guan2023CCD}  & ICCV 2023  & ViT  &  62.1&  55.4&  52.0&\href{https://github.com/TongkunGuan/CCD}{URL} \\
SIGA~\cite{guan2023SIGA}  & CVPR 2023  & ResNet&  69.0&  66.2&  40.4&\href{https://github.com/TongkunGuan/SIGA}{URL} \\
CDistNet~\cite{zheng2024cdistnet}  & IJCV 2023  & ResNet+Transformer  &  66.6&  62.3&  65.5&\href{https://github.com/simplify23/CDistNet?tab=readme-ov-file}{URL} \\
DiG~\cite{yang2022DiG}  & ACM MM 2022  & ViT  &  62.7&  53.2&  52.0&\href{https://github.com/ayumiymk/DiG}{URL} \\
PARSeq~\cite{bautista2022PARseq}  & ECCV 2022  & ViT  &  75.0&  72.7&  23.4&\href{https://github.com/baudm/parseq}{URL} \\
MGP-STR~\cite{wang2022mgpstr}& ECCV 2022& Transformer& 69.6& 67.5& 148.0&\href{https://github.com/AlibabaResearch/AdvancedLiterateMachinery/tree/main/OCR/MGP-STR}{URL}\\
BLIVA~\cite{hu2024bliva} &  AAAI 2024& ViT  &  56.7&  51.3&  7531.3&\href{https://github.com/mlpc-ucsd/BLIVA}{URL} \\
SimC-ESTR~\cite{wang2025eventstr}  &arXiv 2025 & ViT  &  65.1&  56.8&  7531.3&\href{https://github.com/Event-AHU/EventSTR}{URL} \\
\hline 
ESTR-CoT (Ours)  &  -& ViT  &65.6  &  57.1&  7531.3&\href{https://github.com/Event-AHU/ESTR-CoT}{URL} \\
\hline 
\end{tabular}
\end{table*}

\subsection{Dataset and Evaluation Metric}  
We evaluate our method on three event-based scene text recognition benchmark datasets, i.e., \textbf{WordArt*}~\footnote{\url{https://opendatalab.com/OpenDataLab/WordArt}}, \textbf{IC15*}~\footnote{\url{https://aistudio.baidu.com/datasetdetail/96799}}, and \textbf{EventSTR}~\cite{wang2025eventstr}. These datasets are consistent with the evaluation protocol in our previous work~\cite{wang2025eventstr}, allowing for direct comparison and fair benchmarking.

\noindent $\bullet$ \textbf{WordArt*}: Converted from the original WordArt~\cite{xie2022CornerTransformer} dataset using the event simulator ESIM~\cite{rebecq2018esim}, this dataset includes artistic text samples from posters, greeting cards, covers, and handwritten notes. It contains 4,805 training images and 1,511 validation images.
    
\noindent $\bullet$ \textbf{IC15*}: Derived from the ICDAR2015~\cite{karatzas2015icdar} dataset and converted to event-based images. It includes 4,468 training samples and 2,077 test samples from natural scenes.
    
\noindent $\bullet$ \textbf{EventSTR}: Proposed by Wang et al., is a large-scale benchmark dataset for event-based scene text recognition, encompassing diverse character layouts, motion patterns, and illumination conditions. It provides a comprehensive testbed for evaluating both recognition accuracy and temporal robustness, and serves as the foundation for exploring LLM-based reasoning under challenging visual dynamics.

We follow the same evaluation setting as in~\cite{wang2025eventstr}. For EventSTR, we report BLEU-1 to BLEU-4 scores to assess both linguistic accuracy and reasoning quality. BLEU is computed at the character level for Chinese and word level (case-insensitive) for English. For WordArt* and IC15*, we report word-level recognition accuracy.

\subsection{Implementation Details}   
We build our framework upon the BLIVA architecture~\cite{hu2024bliva}, which serves as our vision-language baseline. The model is initialized with pre-trained weights from BLIVA and then fine-tuned on each target dataset individually, allowing it to adapt to domain-specific characteristics in \textit{WordArt*}, \textit{IC15*}, and \textit{EventSTR}.  For optimization, we adopt the AdamW optimizer~\cite{loshchilov2017AdamW} with hyper-parameters $\beta_1 = 0.9$, $\beta_2 = 0.999$, and weight decay of $0.05$. The learning rate is warmed up linearly from $10^{-8}$ to $10^{-5}$ over the first 1,000 steps, followed by a cosine decay schedule that reduces the learning rate to a minimum of $0$. All experiments are conducted on an Nvidia A800 GPU. More implementation details, including prompt construction and loss design, can be found in our source code on GitHub.

\subsection{Comparison on Public Benchmark Datasets} 

\noindent $\bullet$ \textbf{Results on EventSTR.} Table~\ref{tab:bleu} reports BLEU score comparisons on the EventSTR dataset with recent state-of-the-art scene text recognition methods. Traditional approaches such as CCD~\cite{Guan2023CCD}, SIGA~\cite{guan2023SIGA}, and CDistNet~\cite{zheng2024cdistnet} obtain relatively low BLEU-4 scores (<0.31), indicating weak reasoning ability. Transformer-based methods like PARSeq~\cite{bautista2022PARseq} and MGP-STR~\cite{wang2022mgpstr} perform better but are not tailored for event-based inputs. GOT-OCR2.0~\cite{wei2024OCR2.0} achieves a BLEU-4 of 0.332, while SimC-ESTR~\cite{wang2025eventstr} boosts this to 0.430 by incorporating a vision-language backbone. ESTR-CoT consistently improves performance across all BLEU metrics. In particular, it achieves the highest BLEU-1 score of 0.648, matching or surpassing all prior baselines. These results demonstrate the effectiveness of explicit reasoning supervision in event-based scene text recognition.

\noindent $\bullet$ \textbf{Results on WordArt* and IC15* Datasets.}
As shown in Table~\ref{tab:acc}, our method, which leverages pre-training on Visual Question Answering (VQA) data, results in a significant improvement over the baseline. However, the performance on the WordArt* and IC15* datasets is still suboptimal when compared to methods specifically trained on large-scale text recognition datasets like MJ~\cite{jaderberg2014MJ} and ST~\cite{gupta2016ST}. While VQA pre-training helps the model learn visual-textual relationships, it is not fully optimized for complex text recognition tasks, particularly those involving noisy or highly variable backgrounds. These are areas better addressed by models specifically trained on OCR data.

Additionally, the synthetic datasets used for fine-tuning our model (WordArt* and IC15*) have relatively low resolution and lack the complexity and diversity found in larger OCR datasets. This limits the model's ability to reach the performance levels of established methods such as LISTER, CCD, and PARSeq, which benefit from extensive, high-quality OCR training data. Despite this, our approach demonstrates a clear improvement over the baseline, highlighting the potential benefits of VQA pre-training for enhancing visual-textual understanding in OCR tasks. In future work, we plan to explore how large-scale datasets like MJ and ST can be leveraged more efficiently without excessive resource consumption, to further boost performance.

\subsection{Ablation Study}


\noindent $\bullet$ \textbf{Analysis on the Performance of Different Architectures.~} 
\label{sec:architecture_analysis}
In this experiment, we compare the Prompt-based Control architecture and the Projection-separated Control architecture, corresponding to the \textit{diff-prompt} and \textit{diff-projection} variants, respectively. The BLEU scores across different n-gram levels (BLEU-1 to BLEU-4) are reported in Table~\ref{tab:Architectures}.

The Prompt-based Control architecture appends different suffixes (\textcolor{magenta}{<answer>} and \textcolor{blue}{<thinking>}) to the common prompt, guiding the model to generate specific outputs related to the final answer or the reasoning process. This results in a more controlled generation process, where the model is explicitly instructed to focus on different aspects of the task (i.e., the answer or the reasoning). From the table, we can see that this architecture yields slightly better BLEU scores compared to the Projection-separated Control architecture, especially in BLEU-1 and BLEU-2. This suggests that the model benefits from having more explicit guidance through the prompt suffix, improving its ability to generate high-quality answers and reasoning chains.

On the other hand, the Projection-separated Control architecture uses distinct projections for both visual and textual components, allowing each modality to be treated separately before combining them. This separation can help the model handle visual and textual information more effectively but may also lead to challenges in generating a coherent final output when the modalities interact. The slightly lower BLEU scores for this architecture suggest that while it might have advantages in certain visual-textual tasks, the overall output quality could be improved with a more explicit prompt-based control mechanism, as seen in the diff-prompt configuration.

In summary, the slight differences in BLEU scores between these two architectures demonstrate the impact of controlling the generation process. The Prompt-based Control method, by directly influencing the output with suffixes, appears to yield better results for text generation tasks, particularly when reasoning is involved. Future work could explore further refinements in combining both strategies for even higher performance.

\begin{table}
\centering
\small 
\caption{Performance of Different Architectures.} 
\label{tab:Architectures}
\begin{tabular}{c|cccc}
\hline
\textbf{Architectures}&  \textbf{BLEU-1} &  \textbf{BLEU-2}&  \textbf{BLEU-3}& \textbf{BLEU-4}\\ \hline
diff-projection & 0.638 & 0.581 & 0.498 & 0.430\\
diff-prompt   & 0.648& 0.586& 0.500& 0.430\\ 
\hline
\end{tabular}
\end{table}

\newcommand{\yes}{\textcolor{SeaGreen4}{\ding{51}}}
\newcommand{\no}{\textcolor{DarkRed}{\ding{55}}}

\noindent $\bullet$ \textbf{Analysis on CoT filtering.~} In this experiment, we evaluate the effectiveness of our CoT filtering pipeline by comparing the performance of models trained with unfiltered CoT data versus those trained with the filtered CoT data. The models were trained using identical configurations except for the inclusion of the CoT filtering process. Specifically, for the unfiltered case, we removed the automatic evaluation and expert validation steps, directly using raw CoT samples for supervision.

The results, shown in Table \ref{tab:filtering}, indicate a noticeable improvement in the model’s performance after applying CoT filtering. Specifically, the filtered CoT data leads to a higher BLEU score across all n-gram levels (BLEU-1 to BLEU-4). The unfiltered model achieved a BLEU-1 score of 0.632, while the filtered model showed an improvement to 0.648. Similarly, for BLEU-2, BLEU-3, and BLEU-4, the filtered CoT model outperforms the unfiltered model, achieving scores of 0.586, 0.500, and 0.430, respectively, compared to the unfiltered model's 0.574, 0.492, and 0.423.

This demonstrates that the CoT filtering pipeline significantly enhances the quality of the training data, leading to better generalization and more accurate text generation. By removing noisy or irrelevant samples and focusing on high-quality CoT data, the model is able to generate more coherent and contextually accurate reasoning chains, which translates into improved overall performance.

\begin{table}
\centering
\small 
\caption{Performance Comparison with and without CoT Filtering.} 
\label{tab:filtering}
\begin{tabular}{c|cccc}
\hline
\textbf{Filtering}&  \textbf{BLEU-1} &  \textbf{BLEU-2}&  \textbf{BLEU-3}& \textbf{BLEU-4}\\ \hline
\no &  0.632&  0.574&  0.492&0.423\\
\yes & 0.648& 0.586& 0.500& 0.430\\ 
\hline
\end{tabular}
\end{table}

\noindent $\bullet$ \textbf{Analysis on Loss Weighting.~}
\label{sec:Loss_analysis}
We investigate the impact of different loss weighting strategies when jointly training the model to generate both the final answer and the reasoning chain. Specifically, we compare two schemes:

\textit{1). Weighted Loss Combination:} The total loss is a weighted sum of the answer and reasoning losses:
  \[
  \mathcal{L} = \lambda \cdot \mathcal{L}_{\text{answer}} + (1 - \lambda) \cdot \mathcal{L}_{\text{thinking}},
  \]
  where $\lambda \in [0,1]$ controls the relative importance of each objective.
  
\textit{2). Direct Summation:} The total loss is a simple summation of both components without reweighting:
  \[
  \mathcal{L} = \mathcal{L}_{\text{answer}} + \mathcal{L}_{\text{thinking}}.
  \]
  
  

We experiment with different values of $\lambda$ (0.3, 0.5, and 0.7) to examine how emphasizing either the answer or reasoning affects overall performance. As shown in the Table \ref{tab:loss_weighting_comparison}, all weighted schemes perform comparably, with only minor variations across BLEU scores. Notably, the direct summation strategy achieves the highest performance overall. This suggests that giving equal emphasis to both the answer and reasoning generation may encourage better joint optimization and lead to more coherent outputs.

\begin{table}[ht]
\centering
\small
\caption{Performance Comparison of Different Loss Weighting Schemes.}
\label{tab:loss_weighting_comparison}
\begin{tabular}{l|cccc}
\hline
\textbf{Loss Scheme} & \textbf{BLEU-1} & \textbf{BLEU-2} & \textbf{BLEU-3} & \textbf{BLEU-4} \\ \hline
Weighted ($\lambda=0.5$) & 0.638 & 0.581 & 0.498 & 0.430 \\
Weighted ($\lambda=0.7$) & 0.636 & 0.580 & 0.497 & 0.431 \\
Weighted ($\lambda=0.3$) & 0.637 & 0.581 & 0.498 & 0.430 \\
Direct Summation          & \textbf{0.648} & \textbf{0.586} & \textbf{0.500} & \textbf{0.430} \\
\hline
\end{tabular}
\end{table}

\begin{figure*}[!htp]
\centering
\includegraphics[width=\linewidth]{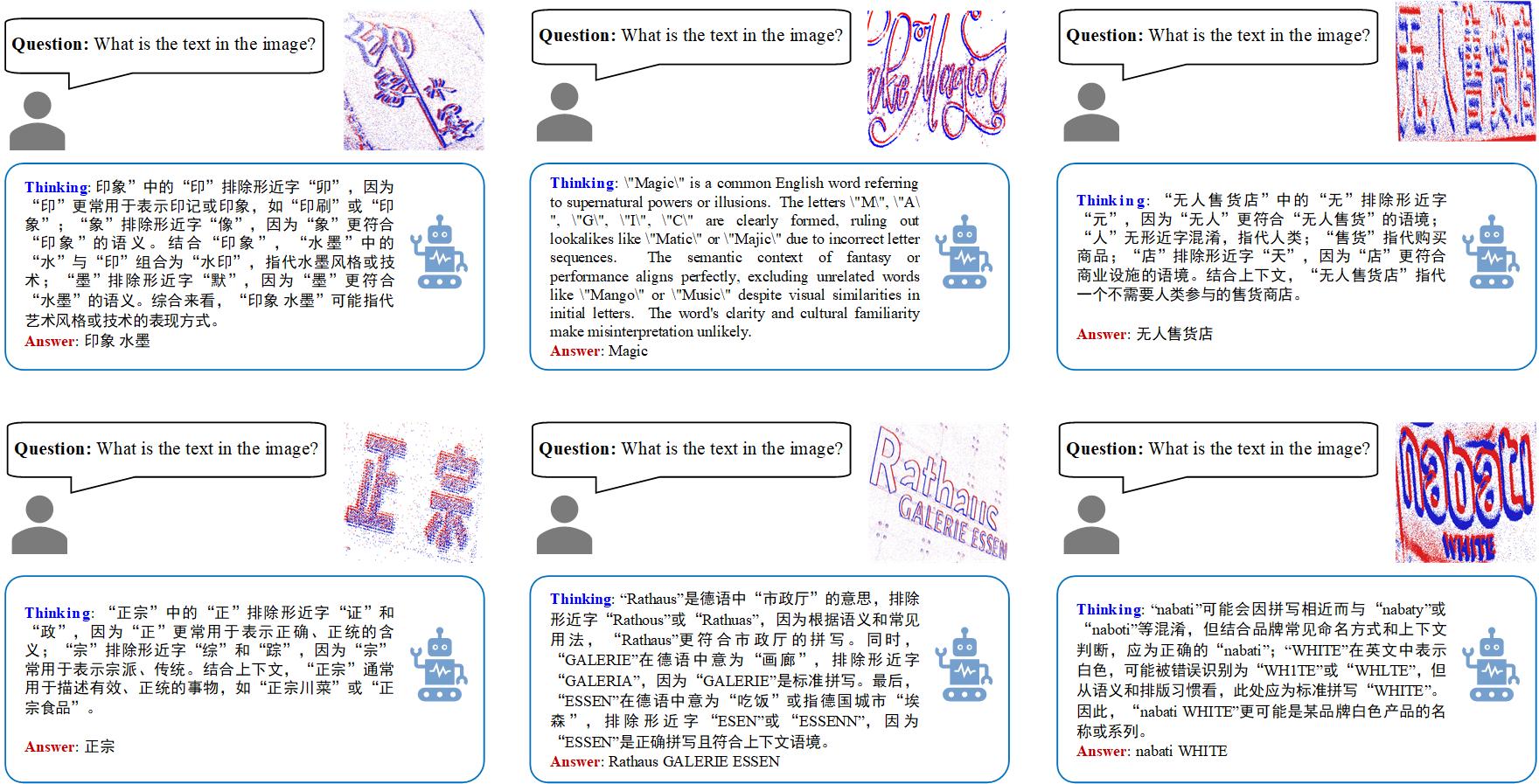}
\caption{\textbf{Visualization of rationales.}
Our model can generate rationales that not only recognize the correct textual content from event-based images but also provide step-by-step reasoning to justify the prediction. These rationales typically include visual disambiguation cues (e.g., excluding visually similar distractors) and semantic justifications (e.g., matching word meaning with context), enabling interpretable and trustworthy scene text recognition under complex or ambiguous visual conditions.}  
\label{fig:reason}
\end{figure*}

\begin{figure*}[!htp]
\centering
\includegraphics[width=\linewidth]{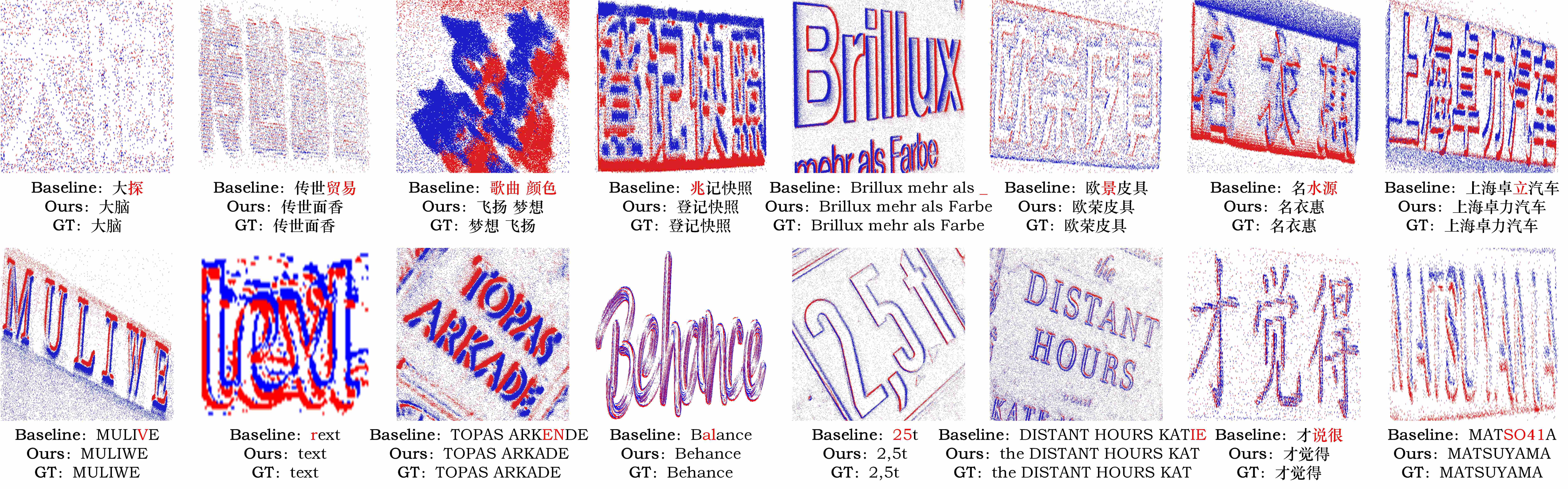}
\caption{\textbf{Comparison of text recognition results: Baseline, Our Proposed Approach, and Ground Truth (GT).}} 
\label{fig:recognition}
\end{figure*}



\subsection{Visualization} 
Fig.~\ref{fig:reason} presents a series of example images processed by our model, accompanied by the corresponding reasoning chains generated for each case. These images showcase various challenging visual-textual scenarios, and the reasoning chains highlight our model’s ability to identify and resolve ambiguities in the text, even in noisy or complex backgrounds. Each image is paired with the question prompt ("What is the text in the image?"), followed by the model’s reasoning process that explains how it arrives at the final output. These examples demonstrate the effectiveness of our model in handling complex text recognition tasks, generating coherent reasoning chains that align with the textual content in the images.

\begin{CJK*}{UTF8}{gbsn}
For instance, in the first image, the model analyzes the text "印象" (impression) and identifies that certain characters can be excluded to clarify the meaning, ultimately recognizing "印象 水墨" (impression ink). Similarly, in the second image, the model distinguishes the word "Magic" from visually similar words like "Matic" or "Majic" based on the context, leading to the correct identification. The reasoning behind these decisions helps the model to generate more accurate final answers, improving both interpretability and performance.
\end{CJK*}

Fig.~\ref{fig:recognition} presents a comparison of text recognition results between the baseline method, our proposed approach, and the ground truth (GT) for a sample image. The input image, along with the recognition results from the baseline, our method, and the GT text, are displayed for comparison. As shown in the figure, the baseline method struggles with accurately identifying the text, especially in challenging conditions such as distorted or noisy text. In contrast, our approach shows a significant improvement in recognition accuracy, demonstrating the effectiveness of our model in handling complex and visually degraded text. The GT text serves as the ideal reference, highlighting the target recognition outcome. This comparison underscores the strengths of our method in achieving better performance on text recognition tasks, particularly in scenarios where the baseline model fails to handle distorted or noisy text.


\subsection{Limitation Analysis} 
While our proposed method significantly enhances the accuracy and interpretability of scene text recognition, there are still some limitations to address:
First, even when configured to output only the final answer without the reasoning chain, our model still suffers from slower inference speed compared to traditional scene text recognition models. This is primarily due to the inherently autoregressive nature and LLM used in our architecture. Such latency can be a drawback in real-time or resource-constrained scenarios.
Second, our model is pre-trained on VQA datasets instead of large-scale OCR-specific corpora (e.g., MJ~\cite{jaderberg2014MJ} and ST\cite{gupta2016ST}). This limits the model’s performance upper bound on standard OCR benchmarks, as domain-specific pre-training has been shown to provide strong inductive biases for text recognition tasks.

\section{Conclusion} \label{sec::conclusion}
To address the limitations of current event stream based scene text recognition  (STR) on the lack interpretability and struggle with contextual logical reasoning, in this paper, we proposed a novel chain-of-thought reasoning-based framework for event stream STR, termed ESTR-CoT. Our approach leverages a vision encoder (EVA-CLIP) to extract visual features from the event stream and aligns them with a pre-trained LLM (Vicuna-7B) via a Q-former module. This allows the model to generate not only accurate text predictions but also interpretable reasoning processes in the form of chain-of-thought. The framework is trained end-to-end through supervised fine-tuning. Furthermore, we introduced a large-scale CoT dataset constructed through a three-stage process (i.e., generation, polishing, and expert verification) to support the training of reasoning-capable models. This dataset lays a solid foundation for future development in this area. Extensive experiments on three benchmark datasets, i.e., EventSTR , WordArt* , and IC15*, fully demonstrate that ESTR-CoT achieves strong performance while significantly improving model interpretability and transparency. 

In our future works, we will further attempt to generate more high-quality reasoning datasets using reinforcement learning technique, such as GRPO (Group Relative Policy Optimization). Also, we will explore how to leverage large-scale OCR datasets (e.g., MJ and ST) in a resource-efficient manner, enabling better recognition performance without significantly increasing computational cost.


\small{ 
\bibliographystyle{IEEEtran}
\bibliography{reference}
}

\end{document}